\documentclass[11pt]{article}
\usepackage[T1]{fontenc}
\usepackage[utf8]{inputenc}
\usepackage{lmodern}
\usepackage[margin=0.96in]{geometry}
\usepackage{amsmath,amssymb,amsthm,mathtools}
\usepackage{booktabs,tabularx,array}
\usepackage{enumitem,microtype,xspace}
\usepackage{pifont}
\usepackage{xcolor,graphicx}
\usepackage[most]{tcolorbox}
\usepackage{algorithm,algpseudocode}
\usepackage{caption,subcaption,placeins,float,wrapfig}
\usepackage{url}
\usepackage{fancybox}
\definecolor{EulerBlue}{HTML}{254C70}
\definecolor{EulerGreen}{HTML}{376450}
\definecolor{EulerGray}{HTML}{F2F4F6}
\usepackage[colorlinks=true,linkcolor=EulerBlue,citecolor=EulerBlue,urlcolor=EulerBlue]{hyperref}
\usepackage[nameinlink,capitalise,noabbrev]{cleveref}
\newcommand{\Euler}{\textsc{Euler}\xspace}
\newcommand{\cmark}{\textcolor{EulerGreen}{\ding{51}}}
\newcommand{\xmark}{\textcolor{red!70!black}{\ding{55}}}
\newcommand{\Fact}{\mathcal G}
\newcommand{\Research}{\mathcal R}
\newcommand{\Memory}{\mathcal M}
\newcommand{\Frontier}{\mathcal F}

\newcommand{\code}[1]{\texttt{#1}}
\newcommand{\state}[1]{\textsc{#1}}
\newcommand{\arxivref}[1]{\href{https://arxiv.org/abs/#1}{arXiv:#1}}
\newcolumntype{Y}{>{\raggedright\arraybackslash}X}
\newcolumntype{L}[1]{>{\raggedright\arraybackslash}p{#1}}
\newtheorem{definition}{Definition}
\setlist[itemize]{leftmargin=1.5em,itemsep=0.1em,topsep=0.25em}
\setlist[enumerate]{leftmargin=1.5em,itemsep=0.15em,topsep=0.25em}
\hypersetup{pdftitle={Euler: Towards Human-Centered Multi-Agent Mathematical Discovery},pdfauthor={Qiuyu Chen, Qi Wang, Liang Xu, Xin Jin, Baao Xie, Wenjun Zeng}}
\definecolor{EulerTitleGray}{HTML}{F4F4F4}
\definecolor{EulerFrameGray}{HTML}{E6E6E6}
\usepackage{titlesec}
\titlespacing*{\section}{0pt}{1.8ex plus .5ex minus .2ex}{.8ex plus .2ex}
\titlespacing*{\subsection}{0pt}{1.3ex plus .3ex minus .1ex}{.5ex plus .1ex}
\begin{document}

\newlength{\EulerTitleBoxWidth}
\setlength{\EulerTitleBoxWidth}{0.92\textwidth}
\begin{center}
\begin{tcolorbox}[enhanced,width=\EulerTitleBoxWidth,colback=EulerTitleGray,colframe=EulerFrameGray,boxrule=0pt,arc=15pt,outer arc=15pt,left=14pt,right=14pt,top=12pt,bottom=10pt]
\centering
{\LARGE\bfseries \includegraphics[height=1.18em]{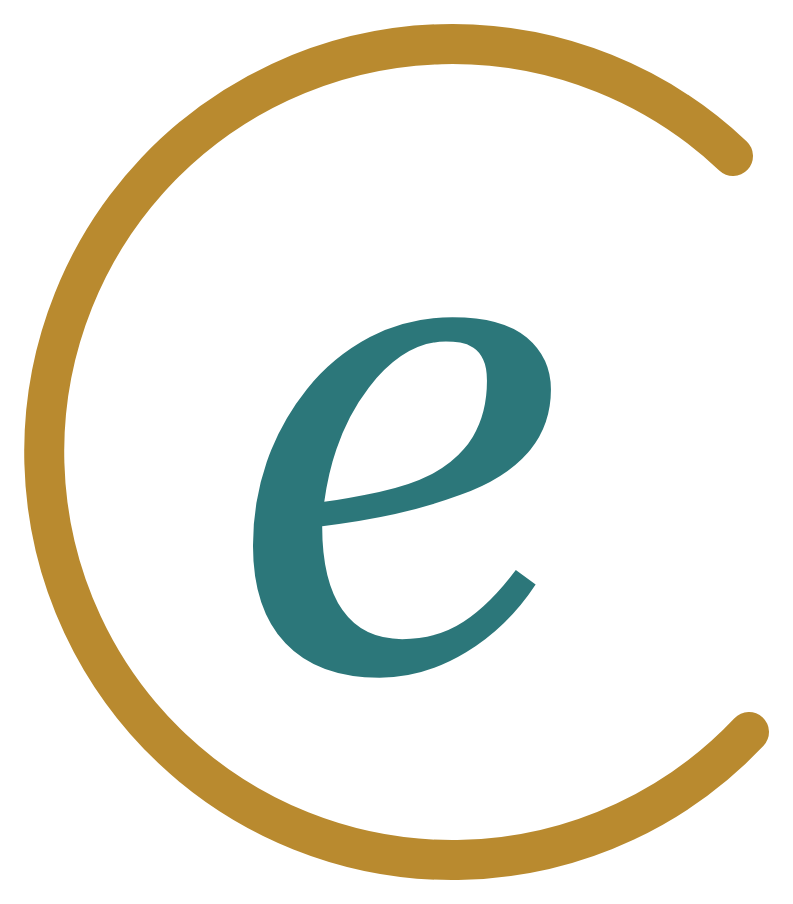}\hspace{0.28em}Euler: Towards Human-Centered\\[0.22em]
Multi-Agent Mathematical Discovery\par}
\vspace{0.80em}
{\normalsize \textbf{Qiuyu Chen}\textsuperscript{1,4}\textbf{,} \textbf{Qi Wang}\textsuperscript{2}\textbf{,} \textbf{Liang Xu}\textsuperscript{3}\textbf{,} \textbf{Xin Jin}\textsuperscript{4}\\[0.18em]
\textbf{Baao Xie}\textsuperscript{4,*}\textbf{,} \textbf{Wenjun Zeng}\textsuperscript{4,*}\par}
\vspace{0.56em}
{\footnotesize
\textsuperscript{1}Shanghai Jiao Tong University, 
\textsuperscript{2}Institute of Automation, Chinese Academy of Sciences, 
\textsuperscript{3}Zhongguancun Academy, Beijing, China,\\[0.15em]
\textsuperscript{4}Ningbo Key Laboratory of Spatial Intelligence and Digital Derivative,\\[0.10em]
Ningbo Institute of Digital Twin, Eastern Institute of Technology, Zhejiang, China\par}
\vspace{0.62em}
\begin{minipage}{0.965\linewidth}
\small
\noindent
Mathematical research drives scientific discovery, and recent advances in AI for mathematics have shown that language-model agents can solve challenging problems and develop substantial mathematical proofs. Yet much of today's AI-for-mathematics work centers on obtaining a final answer. This emphasis leaves the exploration behind a result difficult for researchers to understand, while long-running investigations face fragmented context and the risk of unsupported claims propagating across reasoning steps. We present \Euler, a human-centered mathematical research system built around a shared, structured representation of the research process. Euler organizes mathematical statements, assumptions, proof dependencies, alternative approaches, failed attempts, and verification evidence into connected research graphs. This representation allows researchers to trace how results are developed and supplies agents with persistent, task-relevant knowledge for planning, collaboration, and verification across sessions. Researchers can also select a graph node linked to a saved team state, revisit the agents' earlier reasoning, and explore alternative directions through continued interaction. We illustrate Euler's capabilities through three mathematical research case studies and describe the implemented system and its software validation.\par
\end{minipage}
\vspace{0.66em}
{\small\textbf{Code:} \href{https://github.com/canghaimeng/Euler}{github.com/canghaimeng/Euler}\par}
{\footnotesize\color{EulerBlue}October 13, 2026 \;\textbullet\; Technical Report \;\textbullet\; Version 0.3.0.dev0\par}
\end{tcolorbox}
\end{center}
\vspace{-0.45em}
\begingroup
\renewcommand{\thefootnote}{}
\footnotetext{\textsuperscript{*}Corresponding authors.}
\addtocounter{footnote}{-1}
\endgroup

\section{Introduction}
\label{sec:intro}
Mathematical research provides the conceptual and theoretical foundations of advances across science and technology. A mathematical result can explain a phenomenon, establish the limits of a method, or open a new line of inquiry. Recent progress in AI for mathematics has substantially expanded the role of language models in this process. Mathematical agents can develop extended proofs, search for counterexamples, use symbolic and formal tools, and coordinate specialized reasoning tasks \cite{rethlas2026,danus2026,mmat2026,magenta2026,leap2026}. These developments suggest a broader role for AI in mathematical discovery and in the day-to-day work of mathematicians.

Much of the current work, however, is organized around the production of a final theorem, proof, or counterexample. Mathematical research also yields valuable knowledge along the way: why a certain approach was attempted, which assumptions turned out to be essential, how partial results fit together, and where a promising line of reasoning failed. Understanding these decisions helps a researcher evaluate a proof, identify its most important ideas, and formulate better questions. This is especially relevant to difficult problems that require sustained exploration and may remain unresolved after substantial effort. An AI research system should make this evolving process accessible so that mathematicians can examine and build on the investigation itself.

Long-horizon reasoning introduces a second, closely related challenge. A research campaign may span many model calls, agent conversations, and tool executions. Relevant definitions and earlier proofs can fall outside the current context window; an agent may repeat an unsuccessful approach or overlook the scope of a previous result. More seriously, an unsupported intermediate claim can be carried into later arguments and gradually acquire the appearance of an established premise. These issues become harder to manage as tasks are divided among agents and resumed across sessions. Reliable continuation therefore depends on retaining the mathematical relationships among prior work, selecting the context relevant to each new task, and checking the evidence used to advance the research.

We develop \Euler around a shared, structured representation of mathematical research. The representation records exact statements and assumptions, dependencies among claims, alternative proof strategies, unsuccessful attempts, and the evidence supporting each result. It serves two complementary purposes. For researchers, it provides an inspectable account of how an investigation develops. For agents, it provides persistent mathematical context and explicit targets for planning, collaboration, and verification. Both interact with the same research records as the investigation evolves.

\Euler combines this representation with a multi-agent research workflow. Its research graph organizes open problems and proposed approaches, while a separate graph of admitted results records claims accepted through the configured verification process, building on the Fact Graph architecture of Danus \cite{danus2026}. During execution, agents use the current unresolved problems and their dependencies to select tasks, retrieve relevant prior results and failures, and coordinate further reasoning. Intermediate conclusions enter the shared record with their statements, scope, and checking evidence. The system can consequently continue an investigation across context windows while maintaining a traceable basis for later steps.

The same structure supports human understanding and direct participation. Euler's workbench presents a hierarchical view of the research, from the main problem to individual claims, proof steps, and evidence. A researcher can follow the premises of a conclusion, inspect an abandoned route, or ask agents about a specific mathematical step. Saved team states are associated with research milestones; from a graph node, the researcher can select an associated saved state and resume the corresponding agent conversations in an independent branch. This permits retrospective questioning of crucial ideas and further exploration of alternatives while preserving the original research history.

This report contributes (1) a structured mathematical research state that connects statements, dependencies, strategies, failures, and evidence in a form shared by researchers and agents; (2) a state-guided approach to long-horizon agent reasoning, including context selection, task coordination, and verification-aware updates; and (3) a research workbench that supports proof inspection, focused human questions, and continuation from saved multi-agent states. We examine three mathematical investigations in optimization, graph theory, and geometric group theory to illustrate how the system exposes proof structure and research history. We report reproducible software and integration validation of the implemented workflow and examine the recorded mathematical cases through their source-linked visualizations.

\begin{figure}[t]
\centering
\includegraphics[width=\textwidth]{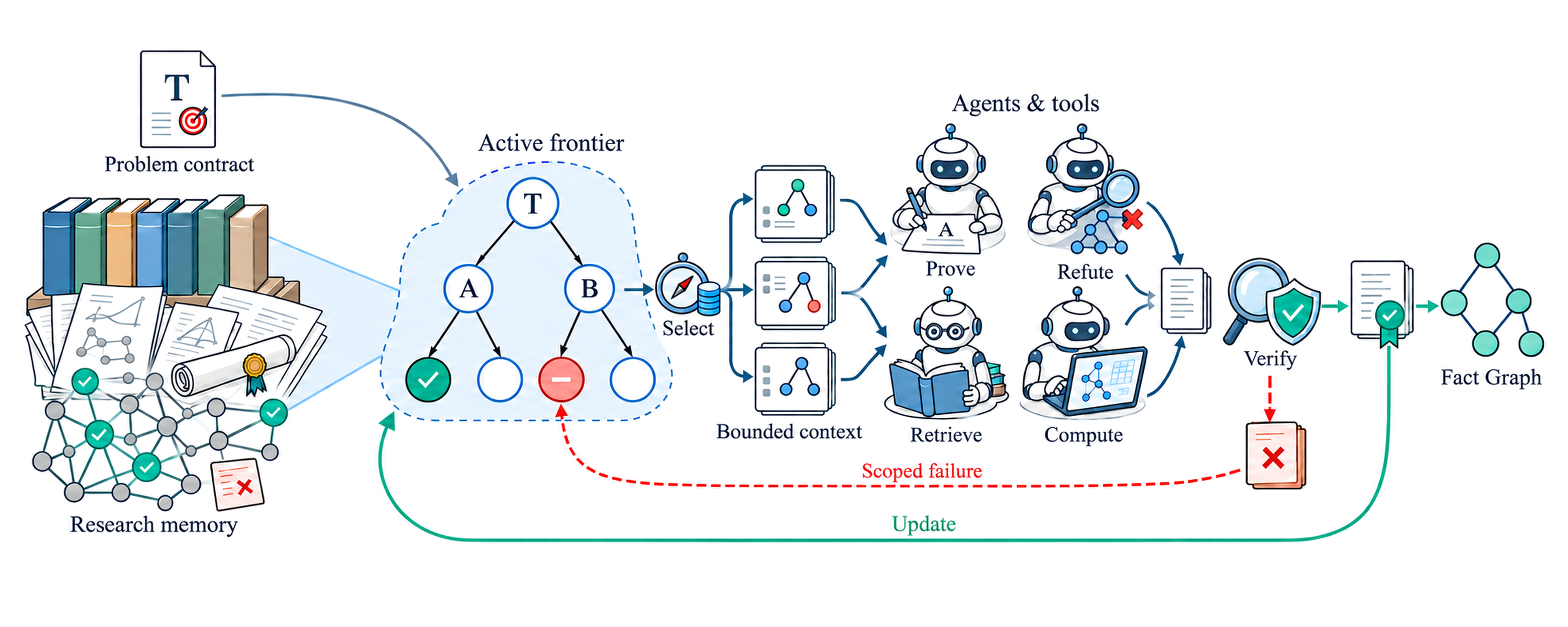}
\caption{The \Euler research loop. A fixed mathematical target and persistent research records determine an active frontier. Agents receive contexts for selected obligations, produce arguments or computational evidence, and return checked outcomes to the shared state. The same underlying objects are presented to the human researcher through the workbench.}
\label{fig:loop}
\end{figure}

\Cref{fig:loop} summarizes the research workflow. 

\section{Research Workflow and System Overview}
\label{sec:requirements}
\subsection{Mathematical context as a persistent research asset}
An extended research project contains several forms of information. Accepted claims are reusable premises. Open obligations describe unfinished work. Routes express proposed ways to connect premises to a target. Experimental and computational artifacts support conjectures or rule out specific constructions. Failed proof attempts identify gaps that may recur when the same mechanism is revisited. These records have different evidential status, yet their relationships determine the next productive step.

\begin{figure}[t]
\centering
\includegraphics[width=\textwidth]{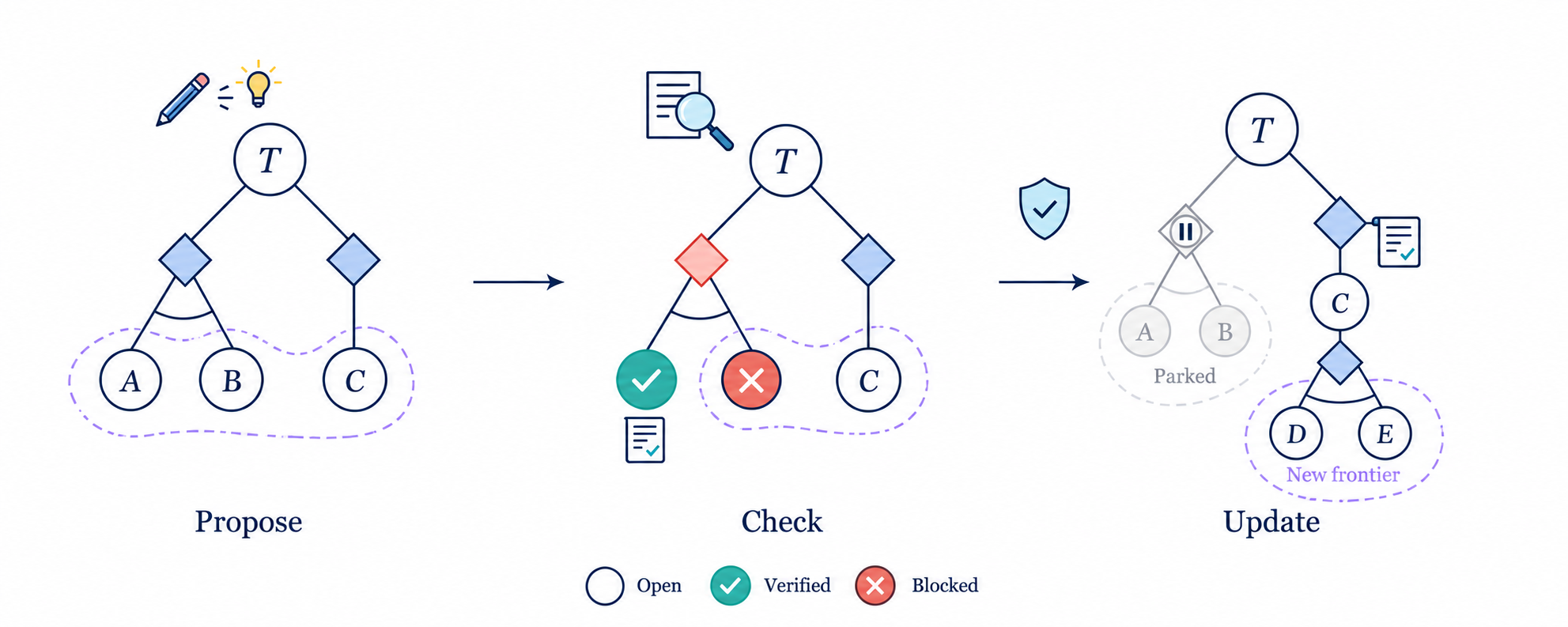}
\caption{An illustrative AND/OR frontier. Two proposed routes compete to establish $T$. Evidence closes $A$ and excludes the route requiring $B$; a checked reduction of $C$ creates obligations $D$ and $E$. The active fringe moves as supported claims and route relations change.}
\label{fig:frontier}
\end{figure}

Consider two candidate routes for a target theorem $T$: one reduces $T$ to $A\land B$, and the other to $C$. A proof of $A$ contributes to the first route, even while $B$ remains open. A checked counterexample to a proposed universal statement $B$ may invalidate that particular route. If a further reduction expresses $C$ in terms of $D\land E$, the active agenda changes, although the number of open obligations may increase. This example motivates a representation in which the progress of an investigation can be assessed through dependencies and checked state changes. \Cref{fig:frontier} displays the corresponding evolution.

The shared representation also supplies a common reference system for human--agent communication. A user question such as ``Does the argument for $B$ require the extra regularity assumption?'' can be attached to the version of $B$ under discussion. The source statement, its assumptions, and its known evidence can then be included automatically in the resulting task. On completion, the researcher can inspect the answer at the same object.

\subsection{System requirements}
These observations lead to four design requirements. \emph{Precise mathematical identity} binds statements to assumptions, quantifiers, domain restrictions, and definition versions. \emph{Evidence-linked explanations} make each accepted conclusion and unresolved dependency inspectable. \emph{Actionable research state} supports targeted retrieval, scheduling, and updates rather than requiring repeated interpretation of full transcripts. \emph{Durable human participation} lets researchers review the state, formulate scoped follow-up work, and recover prior research milestones with their team context.

\subsection{System architecture}
\label{sec:system-overview}
Euler is organized around a persistent mathematical research state that is accessible to both
researchers and AI agents. The \emph{research storage layer} maintains private worker notes,
project-shared findings, admitted mathematical facts, and the graph of active research goals and
approaches. A \emph{research-control layer} selects mathematical tasks, assembles their relevant
context, manages execution budgets, and updates the state when checked results arrive. The
\emph{agent execution layer} connects the lead and its specialized teammates to mathematical
reasoning, theorem retrieval, computation, and verification tools. The \emph{research workbench}
reads the same state to display proof dependencies, active exploration, evidence, and team
conversations, and to prepare focused follow-up work or branch from a saved team state.

A project begins when the researcher specifies a mathematical target and a team. The lead
introduces candidate approaches and assigns work against open goals. Each agent receives a
bounded task context assembled from relevant statements, dependencies, earlier attempts, and
evidence. Proposed results pass through the corresponding checker and are linked back to the
research graph. Throughout the investigation, the workbench exposes the evolving proof structure
and allows the researcher to inspect a claim or continue a specific line of reasoning. This
architecture makes the research representation the shared interface between mathematical
understanding and sustained agent execution, as depicted in \cref{fig:loop}.

\paragraph{Relationship to Danus.}
Euler retains the verifier-gated Fact Graph and three-level memory foundation of Danus
\cite{danus2026}. Its additional mechanisms organize the unfinished mathematical investigation
and connect that representation to execution and human participation. \Cref{tab:danus-compare}
summarizes the publicly documented architecture of both systems.

\begin{table}[!htbp]
\centering\footnotesize
\caption{Published architecture features of Danus and Euler.}
\label{tab:danus-compare}
\setlength{\tabcolsep}{5pt}
\renewcommand{\arraystretch}{1.12}
\begin{tabularx}{\textwidth}{@{}L{0.22\textwidth}Y Y@{}}
\toprule
Capability & Danus \cite{danus2026} & Euler (Ours) \\\midrule
Verified result memory & \cmark\ Fact Graph & \cmark\ Fact Graph + evidence links \\
Open research state & \xmark\ Not explicit & \cmark\ Goals, routes, blockers \\
State-guided context & \xmark\ Not documented & \cmark\ Scope-aware dependency retrieval \\
Human proof inspection & \xmark\ Dashboard / reports & \cmark\ Proof reader + graph views \\
Branchable continuation & \xmark\ No public team-state recovery & \cmark\ Checkpoints + native session forks \\
Agent coverage & \xmark\ Codex-centered team & \cmark\ Multi-provider adapters \\
\bottomrule
\end{tabularx}
\end{table}

\FloatBarrier

\section{Structured Mathematical Research State}
\label{sec:state}
\subsection{Problem contracts and mathematical scope}
Each project begins with a \emph{ProblemContract} $\mathcal P$. It records the target statement $T$, its mathematical scope $\Gamma$, the target kind, source and definition versions, and the required evidence level. The scope includes applicable assumptions, quantified variables, domains, and object restrictions. A contract defines the proposition against which the investigation is evaluated. Research directions that modify these conditions are recorded under derived contracts with explicit links to their parent problem.

The distinction is important when an intermediate claim uses stronger assumptions. For instance, a proof of $T(u)$ at $u=2$ supplies evidence for that parameter value. A statement quantified over an interval requires a separate obligation and corresponding evidence. Similarly, the logical forms $\forall u\,\exists x\,P(u,x)$ and $\exists x\,\forall u\,P(u,x)$ describe different research targets. Versioned scope allows the system to retain these distinctions through task generation and verification.

An obligation $o$ stores an exact statement $q_o$, a contract reference, a mathematical scope $\Gamma_o$, provenance, an evidence requirement, and its current status. Its identity is derived from stable semantic fields, while operational status is maintained through recorded transitions. This permits an obligation to be reopened after evidence changes without losing the earlier history.

\subsection{Persistent memory and the two research graphs}
\Euler uses three memory tiers. Worker-local memory contains private notes and execution records. Shared global memory contains typed findings, including proof attempts, examples, counterexamples, objections, dead ends, and guidance. The \emph{Fact Graph} $\Fact_t$ stores statements and proof texts admitted by the configured verification pathway. Facts include predecessor IDs, glossary information, and external references; their content-addressed identifiers support stable lookup and dependency-based revocation. The Fact Graph and its admission gateway are developed from the Danus architecture \cite{danus2026}.

The \emph{Research Graph} $\Research_t$ describes the wider investigation. Its objects include obligations, candidate routes, evidence records, and typed relations. A route $r$ identifies a parent goal $q_r$, a collection of premise obligations $\operatorname{Prem}(r)$, and a proposed closing argument. A proof route has the form
\begin{equation}
\label{eq:route}
\Gamma_r\;\vdash\;\left(\bigwedge_{o\in\operatorname{Prem}(r)}q_o\right)\Longrightarrow q_r.
\end{equation}
The closing implication is itself subject to an evidence requirement. Multiple routes attached to one goal represent alternative approaches; the premises within each route form conjunctive requirements. Supported reduction relations, equivalences under stated assumptions, refutations, and descent arguments therefore have explicit mathematical and evidential scope.

The implemented relation vocabulary includes \code{REQUIRES}, \code{REDUCES\_TO}, \code{EQUIVALENT\_UNDER}, \code{SUSPECTED\_SAME}, \code{BLOCKED\_BY}, \code{REFUTED\_BY}, and \code{DESCENDS\_BY}. Relations asserting mathematical consequences require supporting evidence. Exploratory relations preserve the status of a proposed connection until that support is supplied.

A blocker identifies an unresolved gap, failed mechanism, or scoped counterexample together with the route or claim it affects. It carries its source and conditions for reconsideration. For example, an obstruction to a candidate construction may rule out that construction under a specified parameter regime. Recording this conclusion with its scope makes the information reusable when a later worker considers a similar route.

\subsection{The active research frontier}
\begin{definition}[Active research frontier]
\label{def:frontier}
For a project contract $\mathcal P$ and persistent research state at time $t$, the active frontier is the projection
\begin{equation}
\label{eq:frontier}
\Frontier_t=\Phi_{\mathcal P}(\Research_t,\Fact_t)
=(O_t,R_t,B_t,E_t),
\end{equation}
where $O_t$ contains unresolved obligations requiring attention, $R_t$ contains currently considered routes, $B_t$ contains applicable blockers, and $E_t$ references evidence relevant to those objects. The inherited graph edges preserve their dependency and source relationships.
\end{definition}
The frontier concentrates the current agenda without discarding the earlier research record. It can be reconstructed from persistent state after an interruption and can change when an obligation is discharged, a route is revised, a blocker is resolved, or new evidence changes the applicability of an earlier claim. Its structure supports both the workbench's main-line and frontier views and the controller's generation of candidate actions.

A newly initialized project may have one root obligation and no proposed routes. The lead agent adds possible reductions and registers their premise obligations. A verified intermediate statement may close an obligation, while an objection to a reduction can return its route for revision. An archived route retains its original argument and supporting artifacts, allowing it to be reopened after a new fact, tool, or assumption changes the situation.

\subsection{Evidence and state transitions}
\label{sec:evidence}
Evidence is represented by typed artifacts associated with exact claims, scopes, contracts, sources, and checker outcomes. The current implementation defines heuristic observations, exact finite results, checked counterexamples, LLM-checked proofs, certificate-checked results, and Lean-kernel-checked proofs. The evidence requirement is assigned when the problem or obligation is created. Admissibility depends on a compatibility relation among evidence kinds and on validity, scope alignment, provenance, and checker outcome.

For an obligation $o$ and a candidate artifact $e$, we write the admissibility predicate as
\begin{equation}
\label{eq:admit}
\operatorname{Admissible}(e,o)
=\operatorname{Valid}(e)\land
\operatorname{ScopeMatch}(e,o)\land
\operatorname{ClaimMatch}(e,o)\land
\operatorname{KindMatch}(e,o)\land
\operatorname{CheckPass}(e),
\end{equation}
where the required kind determines the acceptable checking procedures. Heuristic observations can inform exploration while obligations requiring a checked result retain their stronger completion condition. The implementation further binds evidence to the contract identity and records the status of dependent objects when source evidence is invalidated.

The gateway supports a separate LLM proof-review service. A verdict from this service establishes that an independent model review accepted the submitted proof according to the configured policy. Formal workflows record Lean kernel checks together with attestations relating formal declarations to the intended mathematical statement. Source-backed computational results are checked against their stated domains and, where needed, their realization as original mathematical objects. These distinctions remain available to the reader as evidence metadata.

A state update is applied after a worker action $a_t$ produces an output $y_t$ and associated checks $z_t$:
\begin{equation}
\label{eq:reconcile}
(\Memory_{t+1},p_t)
=\operatorname{Reconcile}(\mathcal P,\Memory_t,a_t,y_t,z_t),
\end{equation}
where $p_t$ is a progress receipt identifying the changed objects and the evidence supporting the change. The controller records obligation discharge, route reduction, scoped refutation, blocker refinement, and operational outcomes through distinct transitions. A final theorem requires an exact root-aligned result and a separate final audit. The dependency cascade is re-evaluated when an accepted supporting fact is revoked.

\begin{table}[htbp]\centering\small
\caption{Evidence-mediated changes to the research state.}
\label{tab:transitions}
\begin{tabularx}{\textwidth}{@{}L{0.25\textwidth}Y Y@{}}\toprule
Result & Recorded support & Frontier effect\\\midrule
Accepted claim & Matching statement, scope, required evidence & Discharge an obligation and update its dependents\\
Validated reduction & Checked closing argument with dependencies & Register a supported route and expose remaining premises\\
Checked counterexample & Witness with an aligned target and scope & Refute the affected local claim or route\\
Verified obstruction & Concrete checker or proof artifact & Refine the blocker and route options\\
New proposal & Source, author, target, and intended mechanism & Add an exploratory route or task\\
Execution outcome & Bound lease, context, and execution receipts & Update runtime status and resource accounting\\\bottomrule
\end{tabularx}
\end{table}

The resulting structure carries two complementary meanings: it is a mathematical record that a researcher can read, and a computational state on which the controller can operate. This shared basis is the central design choice of \Euler.

\section{Frontier-Driven Agent Reasoning}
\label{sec:agents}
\subsection{Planning and execution over mathematical objects}
The research controller converts the current frontier into concrete work. An action identifies its mathematical target, intended operation, expected evidence, selected agent or tool, and resource allowance. The implemented action kinds include proof construction, counterexample search, literature retrieval, computation, verification, and audit. The controller also tracks whether prerequisites are available, whether an equivalent action has already been executed, and whether the proposed mechanism has encountered a relevant obstruction.

A lead agent proposes mathematical directions and delegates work. Workers carry out scoped assignments and publish intermediate findings to shared memory. The persistent record allows the lead to combine results across workers and across separate sessions. When a project contains alternative sufficient conditions, the lead can explore these in parallel. Within a chosen route, independent premise obligations may be assigned separately, while steps whose proof depends on a particular developing argument can be kept within one worker's context.

At each round the controller uses the current research view and resource policy to admit a feasible subset of candidate actions. The scheduler classifies actions by their relationship to the active proof path, readiness, novelty, and expected use of the available resource budget. It returns explicit reasons for selection, deferral, or rejection. Examples include missing or stale dependencies, duplicate actions, recurrence audits, exhausted slots, and budget constraints. The selected action receives a compiled context and an execution lease before the corresponding provider is launched.

\begin{algorithm}[t]
\caption{Frontier-driven research cycle}
\label{alg:cycle}
\begin{algorithmic}[1]
\Require Contract $\mathcal P$; persistent research state $\Memory$; resource policy $\mathbf b$
\State Restore the current research view and unsettled action receipts
\While{the project remains active}
  \State Incorporate completed outputs through evidence-aware reconciliation
  \State Update the frontier $\Frontier\gets\Phi_{\mathcal P}(\Research,\Fact)$
  \If{a root-aligned candidate satisfies the prerequisites for final audit}
    \State Run the configured final audit and record the acceptance decision
    \If{the root obligation is accepted} \State \Return accepted checkpoint \EndIf
  \EndIf
  \State Review repeated or blocked routes; update continuation recommendations
  \State Propose mathematical actions against frontier objects
  \State Select a feasible action set under dependency, slot, and budget policies
  \For{each selected action $a$}
    \State Compile its required mathematical interfaces and bounded historical context
    \State Bind contract, context, allowance, and lease to the worker assignment
    \State Dispatch and persist start/launch receipts
  \EndFor
  \State Await completions or save a resumable checkpoint
\EndWhile
\end{algorithmic}
\end{algorithm}

The controller maintains a distinction between a newly recorded claim, its examination, and the resulting change to the mathematical state. Each action completion is reconciled against the contract, source versions, and recorded evidence. This produces an inspectable connection between an agent's work and the evolution of the research frontier. \Cref{alg:cycle} summarizes the operating cycle implemented by the control and execution modules.

\subsection{Dependency-aware context compilation}
\label{sec:context}
A research project can accumulate substantially more material than a single model context window can hold. \Euler retains the complete record in persistent storage and creates an action-specific \emph{ContextPacket} for each worker. Its construction is driven by the target obligation and the dependencies needed to reason about that obligation.

Let $a$ be a selected action, $D_t(a)$ the necessary dependency interfaces, and $H_t(a)$ an eligible subset of historical findings and blockers. Conceptually, the context compiler implements
\begin{align}
\label{eq:compile}
C_t(a)&=\operatorname{Compile}\bigl(\mathcal P,a,\Frontier_t,D_t(a),H_t(a)\bigr),\\
\operatorname{size}(C_t(a))&\leq L_a,\qquad |H_t(a)|\leq K_a,
\end{align}
where $L_a$ and $K_a$ are the configured context and historical-recall bounds. The compiler preserves mandatory statements, assumptions, proof interfaces, and required evidence references. Optional history is included within the remaining allowance. If the mandatory material exceeds the available limit, the controller reports an explicit context failure or deferral.

Dependency recall follows the graph of admitted results and unresolved obligations associated with the action. An accepted theorem can be referenced through its statement, hypotheses, and identifier. The full proof remains retrievable for detailed inspection or auditing. Context construction validates required predecessors against their current versions and validity status. This reduces the need to reproduce a large transitive sequence of proof texts in every worker prompt while retaining access to the derivations on which the task relies.

Historical recall supplies nearby failed attempts, counterexamples, obstacles, and earlier verification objections. Retrieval uses explicit target and branch links followed by bounded lexical search and scope-aware selection. An earlier failed proof of a premise may therefore be useful when a new worker attempts that premise using a modified technique; a counterexample outside the task's parameter regime is retained as historical information with its original conditions. The context packet includes a manifest of selected and omitted material, allowing execution records to identify what the worker received.

The compiler places prerequisite assumptions and previously admitted theorem interfaces in the worker's task context while preserving full proofs for inspection. Version checks and the omission manifest make the supplied material explicit and traceable to the persistent research state.

\subsection{Reusing unsuccessful work and detecting recurrence}
Unsuccessful investigations provide information about the next useful experiment. \Euler stores typed findings such as attempted reductions, verifier objections, counterexamples, and dead ends, with links to the relevant branch and subgoal. The execution layer can group repeated failure signatures and request a synthesis of the shared obstruction. The resulting summary remains a research record with its associated provenance and scope.

The control layer also examines recurrence within the Research Graph. A route can return to a previously encountered obligation even after its notation or auxiliary definitions have changed. Recurrence analysis compares graph structure and the versions of applicable premises, evidence, and tools. Strongly connected components identify closed loops in the projected dependency structure, and the controller examines whether those loops contain valid descent relations or alternative exits.

For example, an induction argument for $P(n)$ may reduce the step to $P(n-1)$ through a justified implication and a base case. The dependency relation revisits the same predicate schema while decreasing a well-founded parameter. The research state records such descent evidence explicitly. When repeated actions return to an unchanged obligation without a comparable justification or new evidence, the policy can request an audit, test a counterexample, revise the route, or park the approach until its reopening conditions are met.

Recurrence is therefore evaluated at the level of an affected mathematical route. A new fact unrelated to that route leaves its outstanding blocker unchanged. In contrast, a newly established bridge lemma can change the viability of multiple routes at once. The retained graph supplies the scope and dependency information needed for this distinction.

\subsection{Verification, computational evidence, and final acceptance}
\label{sec:checking}
Mathematical evaluation is integrated into execution at several levels. Proposed natural-language proofs pass through the independent LLM verification gateway when their contract permits that evidence kind. \Euler also supports checked finite computations and a Lean integration that runs a pinned project and records the kernel-check result, source hashes, selected declarations, and permitted axioms. An alignment attestation binds the formal declaration to the intended natural-language statement. The evidence requirements of the corresponding obligation determine whether a checked artifact is sufficient for discharge.

The current built-in compute service includes a bounded integer-polynomial counterexample workflow, with a separate exact checker and an original-object realization review. Agents may invoke additional mathematical tools through their configured environments. Exact computations, incomplete searches, timeouts, and tool failures are recorded with their separate outcome types and scopes. This information is available both to the controller and to the human reader.

At the project level, acceptance involves the original contract. A successful component check closes the corresponding local obligation. A root-level acceptance requires the matching root claim, supporting evidence at the required level, and a distinct final audit. A checked refutation must satisfy the problem's original assumptions and, when the counterexample is formulated in an auxiliary model, its mapping to the original mathematical object. The control plane records terminal outcomes such as \state{Solved} or \state{Refuted} separately from resumable operational states such as budget exhaustion, frontier stagnation, missing external input, and infrastructure failure.

\subsection{Resource accounting and long-horizon continuation}
Tasks are admitted against resource policies that account for monetary cost, reported tokens, wall time, compute use, and concurrency. Let $\mathbf u_t$ denote settled additive resource use, $\boldsymbol\ell_t$ the outstanding reservations, $\widehat{\mathbf c}(a)$ the allowance of action $a$, and $\mathbf b$ the resource envelope. The additive dimensions satisfy
\begin{equation}
\label{eq:budget}
\mathbf u_t+\boldsymbol\ell_t+
\sum_{a\in S_t}\widehat{\mathbf c}(a)\preceq\mathbf b,
\end{equation}
for an admitted batch $S_t$. Deadlines and simultaneous execution slots are checked separately. Reported usage can be incomplete, so missing quantities are preserved as unknown in the accounting record.

The scheduler allocates effort among load-bearing obligations, low-cost discriminating tests, verification, and exploratory routes. Resource reservations also preserve capacity for the final audit. A research run can pause with its graph and pending obligations intact and resume under an updated user allowance. This enables long investigations to retain their mathematical history without requiring an uninterrupted agent session.

\section{Human Understanding and Research Continuation}
\label{sec:human}
\subsection{Reading a mathematical investigation}
The workbench is a visual projection of the shared research state. Its entry point is the original problem contract and the current status of the main goal. The overview presents a compact proof structure and the active routes leading to that goal. The Proof Reader expands individual records into their exact statements, direct premises, arguments, checking artifacts, and later dependent claims. The Relation Map exposes upstream and downstream connections. Live Research presents active assignments and recent investigations; Diagnostics displays execution and control events; Team Interaction provides persistent conversations with registered agents.

The interface supports a progression from overview to mathematical detail. A researcher examining a conclusion can inspect its support recursively and identify where a proof depends on an unproved premise. A researcher reviewing an unsuccessful approach can locate the affected route, read the counterexample or verifier objection, and see how the remaining obligations changed. The same inspection path is available for an open project whose main goal has not yet been settled.

The workbench distinguishes the recorded status of a mathematical object through its evidence type and corresponding source links. The original mathematical text is preserved in the reader, with summaries used for navigation. This is particularly useful in large projects where a small overview module represents several lower-level derivations. Historical case-study records can be presented through a source-bound projection; writable research controls resolve exact current backend identities.

\subsection{Human questions attached to mathematical objects}
\label{sec:interventions}
A researcher may select a current obligation, route, blocker, evidence record, relation, or contract and open its canonical object inspector. The inspector supplies the source object's statement, scope, provenance, and version. The researcher then specifies an intended operation and reviews the resulting task before execution. Implemented intents include verification or audit, challenge and counterexample search, assumption exploration, and alternative-route investigation.

The request binds the project, source object, object version, source contract, exact statement and scope, reviewed derived target, and allowed resources. The control service calculates a preview of the proposed investigation; submission rechecks the binding against the current research revision. For a new hypothesis or conclusion, the system creates a derived problem contract and corresponding open obligation linked to the original source. The researcher must explicitly start the prepared focused action. This design provides a stable mathematical referent throughout the exchange and makes the outcome traceable to the initiating question.

A focused investigation follows the same execution path as a scheduled agent task. The control plane compiles its source context, checks worker and budget policy, issues a lease, and dispatches the action. Its returned evidence is evaluated under the derived contract and reconciled with that action. The UI displays the recorded stages of the interaction, including validation, work creation, execution, and outcome. A single focused request authorizes one bounded action; the user can initiate a subsequent investigation after reviewing its result.

The current control implementation supports pause, resume, and stop at the campaign level. During a pause, new ordinary tasks are withheld, and issued or running leases follow the selected drain or cancellation policy. A separately confirmed focused action can run while the ordinary campaign remains paused. Stop also blocks focused starts. Resume checks outstanding settlements, dependencies, and the current resource allowance. These operations are persisted so the human-readable research history includes the circumstances under which work was authorized and resumed.

\subsection{Recoverable milestones and persistent team sessions}
\label{sec:milestones}
A researcher often needs to revisit an earlier point with the information and team context available at that time. \Euler stores milestones in private local Git history. A checkpoint captures project files, research graphs, evidence artifacts, assignments, receipts, and supported native conversation archives. Its manifest records file hashes and omitted runtime material. Capture occurs at a quiescent boundary, so a milestone corresponds to a consistent recorded team state.

\begin{figure}[!htbp]
\centering
\includegraphics[width=0.98\textwidth]{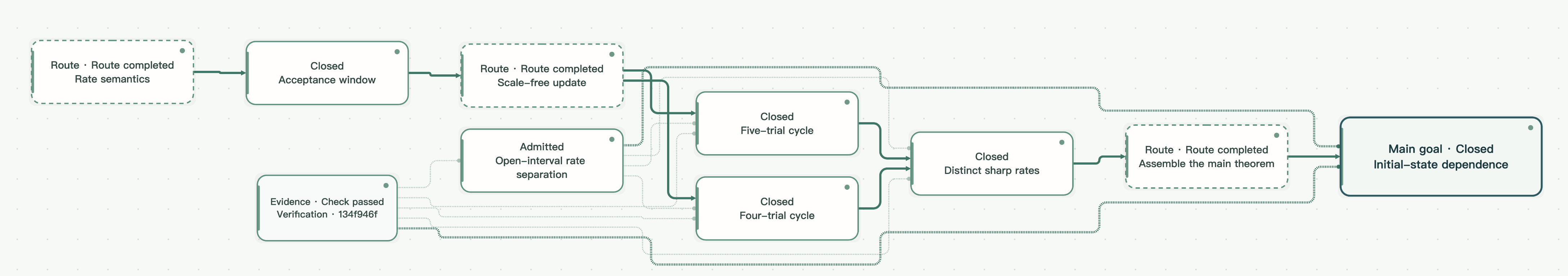}
\caption{Selected dependency chain from the Lewis--Overton research graph. The displayed route runs from rate semantics and scale-free update through the five-trial and four-trial constructions to distinct sharp rates and the final initial-state-dependence result.}
\label{fig:restored-map}
\end{figure}

Restoring a milestone creates an independent workspace. The original investigation remains intact, while the restored branch contains the captured mathematical graph, source files, tasks, and team metadata. The workbench associates each branch with its source commit and permits researchers to navigate its proof structure. A follow-up question can then be posed against the restored research state. \Cref{fig:restored-map} focuses on the main proof dependencies in the Lewis--Overton project.

The native team subsystem extends file recovery to provider sessions. It supports Codex, Claude Code, Gemini CLI, Grok, and Cursor Agent through provider-specific adapters. Depending on the provider, the restored member receives a native session fork or continues a preserved session in a branch-private store. Provider credentials remain outside the milestone archive. The restoration procedure checks availability of the required native-session material before launching a new branch.

\begin{figure}[!htbp]\centering
\includegraphics[width=0.70\textwidth]{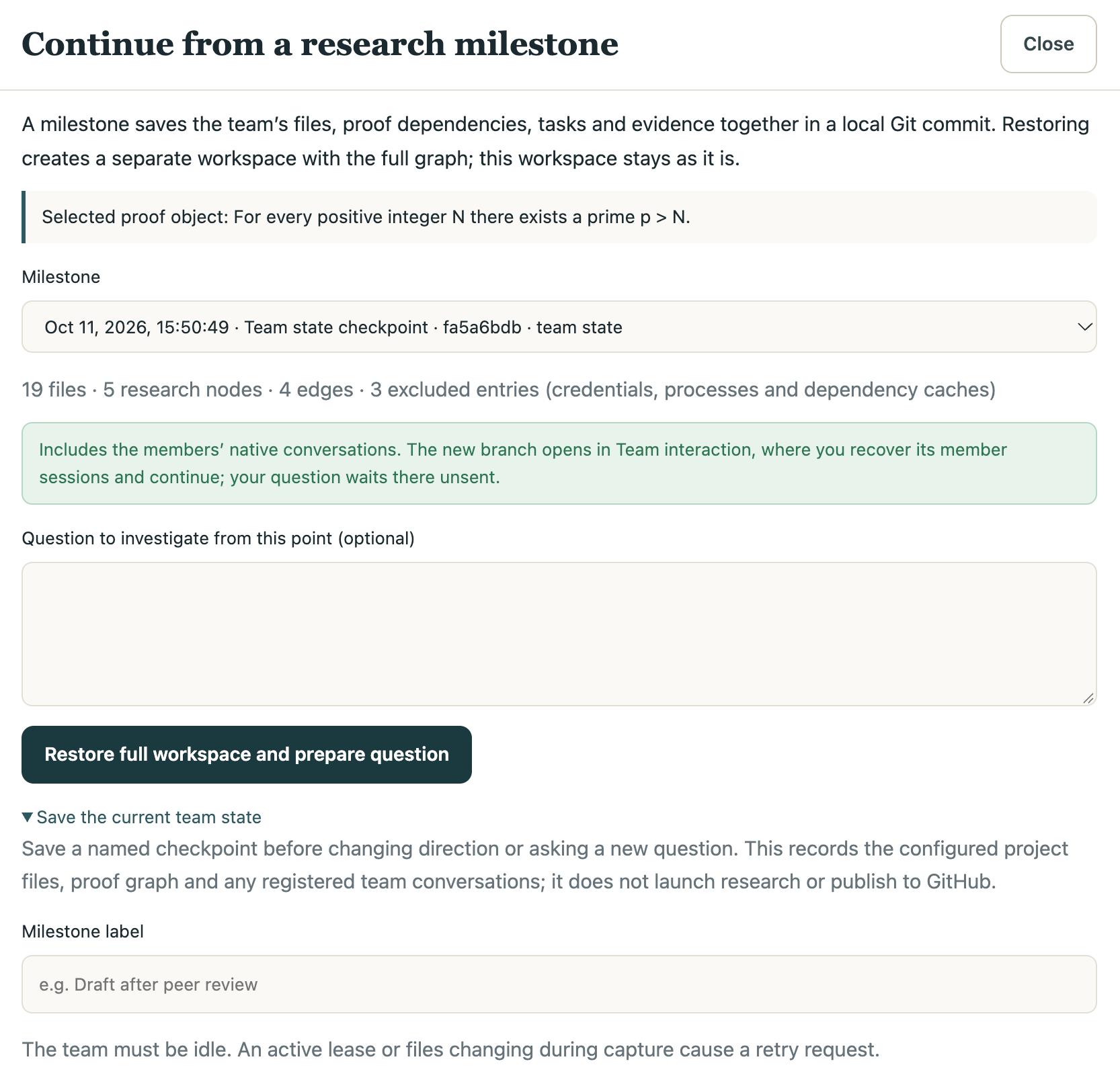}
\caption{Milestone-restoration dialog in the Euclid onboarding exercise. The panel identifies the selected proof object, the saved checkpoint, the recovered branch contents, and the availability of native team conversations. The researcher may attach an optional follow-up question before opening the restored workspace.}
\label{fig:team-room}
\end{figure}

The restoration dialog summarizes the selected proof object, the checkpoint to recover, the contents of the saved branch, and whether the members' native conversations are available for continuation. A researcher can optionally attach a follow-up question before opening the restored workspace. \Cref{fig:team-room} shows this step in the Euclid onboarding exercise; once the workspace is created, the recovered team conversations can be resumed there.

\subsection{From a research question to a traceable outcome}
\begin{figure}[!htbp]\centering
\includegraphics[width=0.92\textwidth]{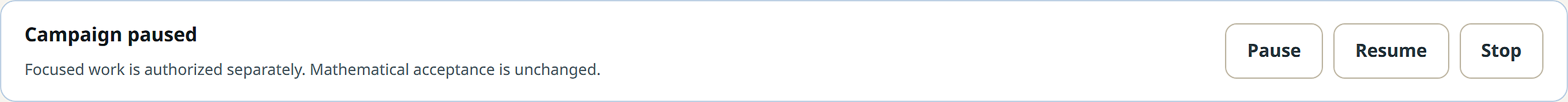}
\caption{Paused parent campaign in the BFGS intervention demonstration. The focused investigation has its own explicit start.}
\label{fig:intervention}
\end{figure}

The workbench links a researcher's question to a particular mathematical object and its source state. \Cref{fig:intervention} shows the paused parent campaign in the contemporary BFGS demonstration. The researcher can inspect the existing graph and initiate a separately authorized, focused investigation while ordinary dispatch is paused.

\begin{figure}[!htbp]\centering
\begingroup
\color{black!43}\setlength{\fboxsep}{0pt}\setlength{\shadowsize}{3pt}
\shadowbox{%
\includegraphics[width=0.77\textwidth]{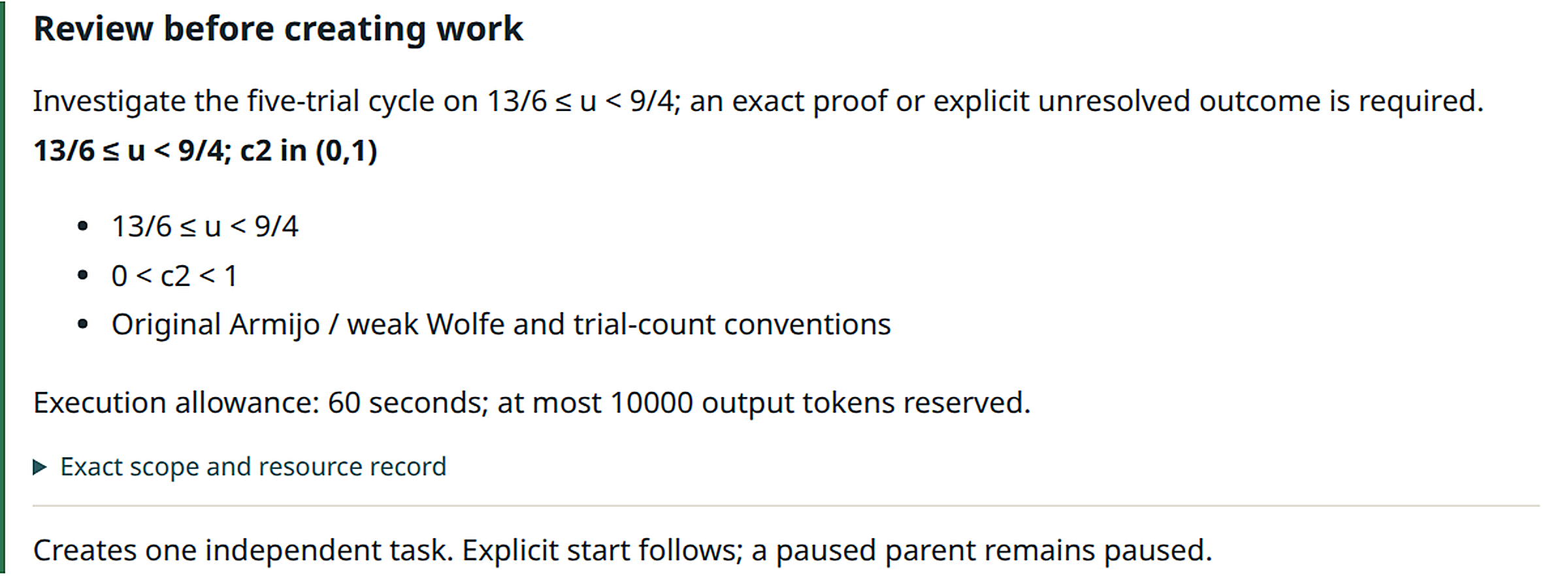}%
}
\endgroup
\caption{Review of the derived BFGS question, assumptions, and execution allowance before dispatch.}
\label{fig:intervention-preview}
\end{figure}

For the proposed follow-up, the researcher reviews the new mathematical statement, parameter scope, and task allowance before confirming the request. \Cref{fig:intervention-preview} shows the reviewed five-trial-cycle question. Its resulting task retains the selected source object's identity and the relevant mathematical context.

\begin{figure}[!htbp]\centering
\begingroup
\color{black!43}\setlength{\fboxsep}{0pt}\setlength{\shadowsize}{3pt}
\shadowbox{%
\includegraphics[width=0.60\textwidth]{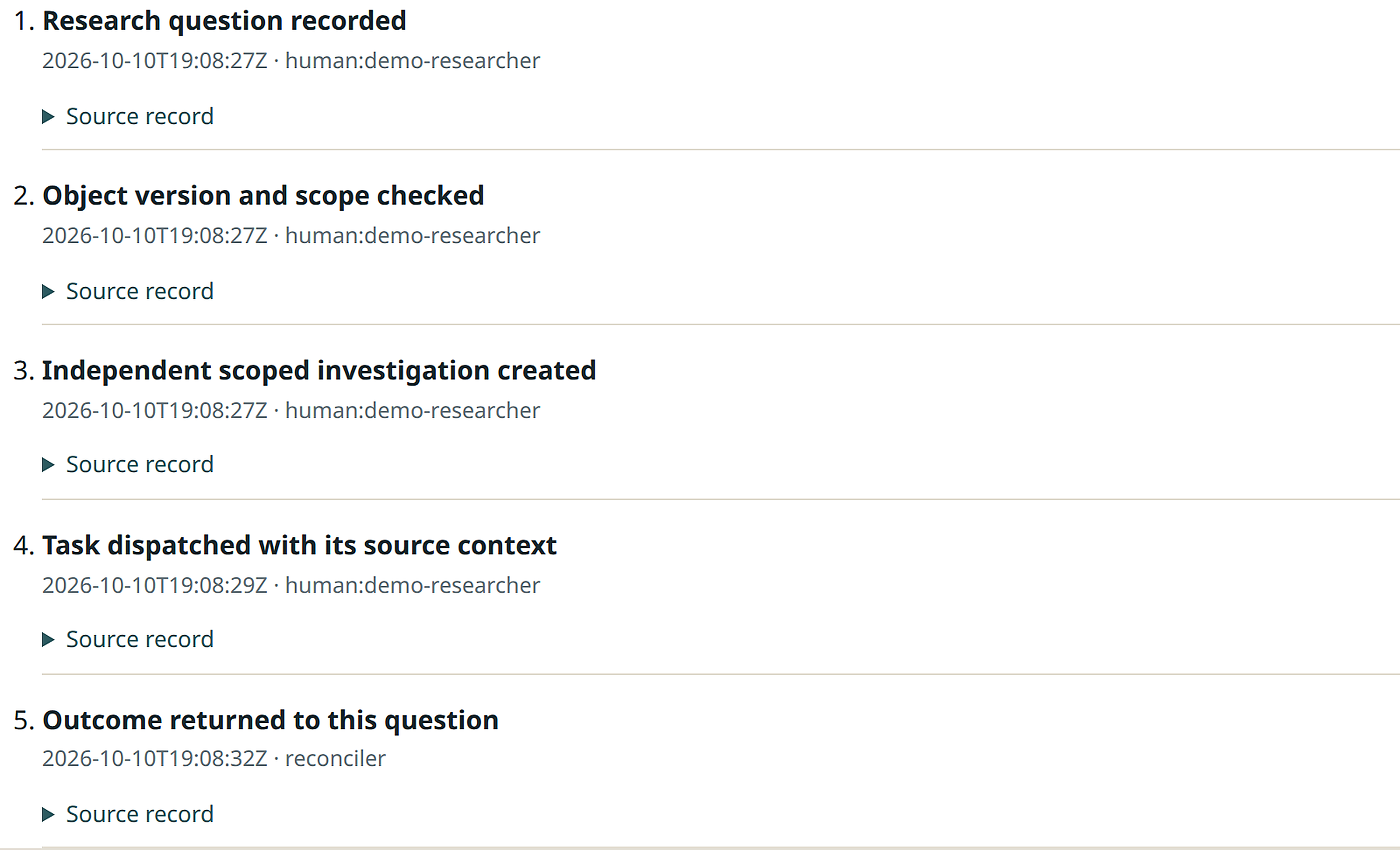}%
}
\endgroup
\caption{Source-linked lifecycle of a focused BFGS investigation, from the researcher's question to its recorded outcome.}
\label{fig:intervention-timeline}
\end{figure}

Once the agent has run, the same request identifier links validation, work creation, dispatch, checking, and reconciliation. The workbench presents these records as the source-linked timeline in \cref{fig:intervention-timeline}. The depicted mathematical worker and verifier responses are deterministic fixtures; the underlying task and record transitions use the implemented execution services.

\FloatBarrier

\section{System Implementation}
\label{sec:implementation}
\subsection{Architecture and persistence}
The current source snapshot is revision \code{ea505c7} of \href{https://github.com/canghaimeng/Euler}{\code{canghaimeng/Euler}}, with the package version \code{0.3.0.dev0}. The Python package separates the mathematical data model, research controller, execution and provider interfaces, verification gateway, native team management, and web observability services. The core mathematical storage layer and the Research Control Plane use standard-library data structures and persistence primitives; external framework dependencies are added for provider access, MCP, verification HTTP services, and the workbench.

The \code{euler.core} package maintains worker-local notes, typed project-shared memory, and the content-addressed Fact Graph. The \code{euler.gateway} package exposes role-gated mathematical tools, including fact submission, retrieval, memory updates, and theorem search. Workers can request fact admission; the coordinating agent receives planning and retrieval capabilities; the verifier receives restricted read access. The verification service supplies an independent review result to the gateway, which records the verdict and proposed mathematical content in the appropriate persistent stores.

The \code{euler.control} package provides the research-state and runtime machinery. Problem contracts, obligations, routes, relations, artifact records, and checking results have explicit schemas. An append-only SQLite event store records state-changing operations with project, actor, causation, idempotency, and payload identities. Research views and runtime views are reconstructed from these records. Scheduler reservations, execution leases, context manifests, and result receipts link proposed work to the concrete action eventually executed. Reconciliation settles a result against these links and updates the corresponding mathematical object when its evidence meets the required conditions.

The execution and provider subsystems run command-line agents through provider-specific adapters. The native team subsystem retains member identities, roles, messages, sessions, and checkpoints. The browser workbench is served by \code{euler.observability} and renders research objects into reading and interaction views. Its front end includes a self-contained reader, graph visualization, team controls, and English/Chinese interface support. \Cref{tab:modules} lists the main implementation components.

\begin{table}[htbp]\centering\small
\caption{Principal software components in the current Euler repository.}
\label{tab:modules}
\begin{tabularx}{\textwidth}{@{}L{0.27\textwidth}L{0.35\textwidth}Y@{}}\toprule
Layer & Source modules & Main responsibility\\\midrule
Mathematical memory & \code{core}, \code{gateway}, \code{verify} & Shared findings, Fact Graph, admission and revocation\\
Research control & \code{control} & Contracts, obligations, frontier, scheduling, context, evidence, final audit\\
Agent execution & \code{execution}, \code{providers}, \code{strategy} & Worker rounds, provider adapters, consultation and execution receipts\\
Team recovery & \code{workspace} & Native sessions, messages, Git milestones and branch isolation\\
Research workbench & \code{observability} & Proof reading, graph navigation, live state, human requests, team conversations\\
Research outputs & \code{write\_paper}, \code{human\_summary} & Source-grounded manuscript and progress-report workflows\\\bottomrule
\end{tabularx}
\end{table}

\subsection{Runtime identity, access, and recovery}
The research controller retains a mathematical object's content-derived identity separately from its changing runtime status. A human request also includes a versioned binding to the active project instance. This permits exact source resolution, replay-safe submission, and detection of stale task contexts. The same mechanism is used when a restored branch refers to an earlier source object. Publicly displayed historical modules remain source-backed reading objects until they are connected to a valid native control identity.

The workbench is designed for a locally controlled research machine or private host. The current personal-workbench deployment uses server-configured project and worker allowlists with same-origin checks; browser interaction does not require a separate control token. Filesystem scope and authorization policy are selected by the host configuration. Native team processes use Linux Landlock or macOS Seatbelt for filesystem isolation; their network capabilities follow host policy. Persistent provider authentication remains on the host, outside the exported public observations and milestone source packages. SSH forwarding supports remote access to a local workbench.

Milestone capture records regular project files, ignored research artifacts, the SQLite event database, and supported native conversation archives at an idle boundary. Restoring a milestone validates the captured content before creating a new workspace and preparing each provider's session according to its supported recovery protocol. Pending work remains suspended until authorized by the researcher. These mechanisms let earlier research states be inspected and continued while preserving branch provenance.

\subsection{Deployment and demonstration entry points}
The repository provides a no-account dashboard demonstration with a fixed elementary proof workflow, an editable personal-research initializer, and commands for configuring agent teams and milestones. A native research workspace specifies its mathematical target, evidence requirement, and member roster, followed by provider preparation and workbench startup. The bundled Euclid exercise demonstrates lead--assistant exchanges and team restoration. A separate human-intervention demonstration exercises object selection, scoped task creation, dispatch, and reconciliation with deterministic model fixtures.

Software tests cover persistence, permission boundaries, schema validation, role-gated tools, context compilation, scheduling, kernel adapters, filesystem isolation, provider protocols, milestone restoration, and browser rendering. These tests form the implementation evidence described in \cref{sec:evaluation}. The mathematical case-study archives, by contrast, are source-bound records of investigations conducted across earlier workflow revisions; their papers and formalization projects are linked directly from the examples directory.

\section{Mathematical Case Studies}
\label{sec:cases}
We examine three mathematical investigations represented in the \Euler workbench: the Nikoghosyan Hamiltonicity conjecture, a Lewis--Overton BFGS rate conjecture, and the four-connected version of the Jankiewicz--Norin--Wise legal-system problem. Each case presents a different type of research structure. The first concerns closing a theorem under exactly its original hypotheses; the second requires precise interpretation of an asymptotic assertion and comparison of two parameterized constructions; the third combines constructive proof with failed candidate searches and a hereditary condition.

These investigations are documented in accompanying manuscripts, publicly linked Lean projects, and historical research-state visualizations. The archived records originate from earlier research workflows, including human guidance and separate formalization activity. We use the workbench to study the organization and inspectability of that mathematical material. \Cref{tab:cases} summarizes the exact questions and their characteristic research structures.

\begin{table}[htbp]\centering\small
\caption{Mathematical scope and representation questions in the three case studies.}
\label{tab:cases}
\begin{tabularx}{\textwidth}{@{}L{0.24\textwidth}Y Y@{}}\toprule
Case & Documented mathematical result & Research-state focus\\\midrule
Nikoghosyan & Hamiltonicity for finite, 1-tough, $(P_4\cup P_1)$-free simple graphs on at least three vertices & Prerequisite closure and exact theorem assumptions\\
Lewis--Overton & Two nonterminating BFGS executions with distinct sharp trial-normalized rates for $9/5<u<13/6$ & Source conventions, paired constructions, and parameter scope\\
Jankiewicz--Norin--Wise & A 33-vertex witness for the four-connected legal-system obstruction question & Candidate exclusions, strategy changes, and hereditary subgraph conditions\\\bottomrule
\end{tabularx}
\end{table}

\subsection{Nikoghosyan: tracing a Hamiltonicity proof to its assumptions}
\label{sec:niko}
A Hamilton cycle visits each vertex of a finite graph exactly once. A graph is 1-tough if, whenever deletion of a vertex set $S$ disconnects the graph into at least two components, the number of components is at most $|S|$. Nikoghosyan conjectured that every 1-tough graph with no induced $P_4\cup P_1$ is Hamiltonian, where $P_4\cup P_1$ denotes a four-vertex path together with an isolated vertex \cite{nikoghosyan2013}. Related work studied the corresponding forbidden-subgraph class under stronger toughness assumptions \cite{shan2025}.

The companion development establishes the asserted Hamiltonicity for finite simple graphs with at least three vertices under the stated 1-toughness and forbidden-subgraph conditions \cite{chennikoghosyan2026}. Its publicly linked Lean repository contains a theorem producing a spanning cycle from edges of the original graph and a corollary expressed using \code{SimpleGraph.IsHamiltonian}. The proof develops the prerequisite factor-existence and minimum-cycle-factor arguments from the theorem's graph hypotheses. The formal source and archived checking record identify the target declarations, build environment, and their transitive axiom dependencies.

From a research-state perspective, the decisive issue is the closure of the prerequisite chain. A proof may develop a useful minimum-factor argument that assumes the existence of a suitable factor; the full target additionally requires the factor to be constructed from the original hypotheses. Likewise, a structural cut or local-restoration lemma can be used in the final argument only after its own conditions have been established. The Research Graph expresses these as separate obligations. The main theorem's dependency closure thereby becomes a concrete object for inspection and review.

\begin{figure}[!htbp]\centering
\includegraphics[width=0.91\textwidth]{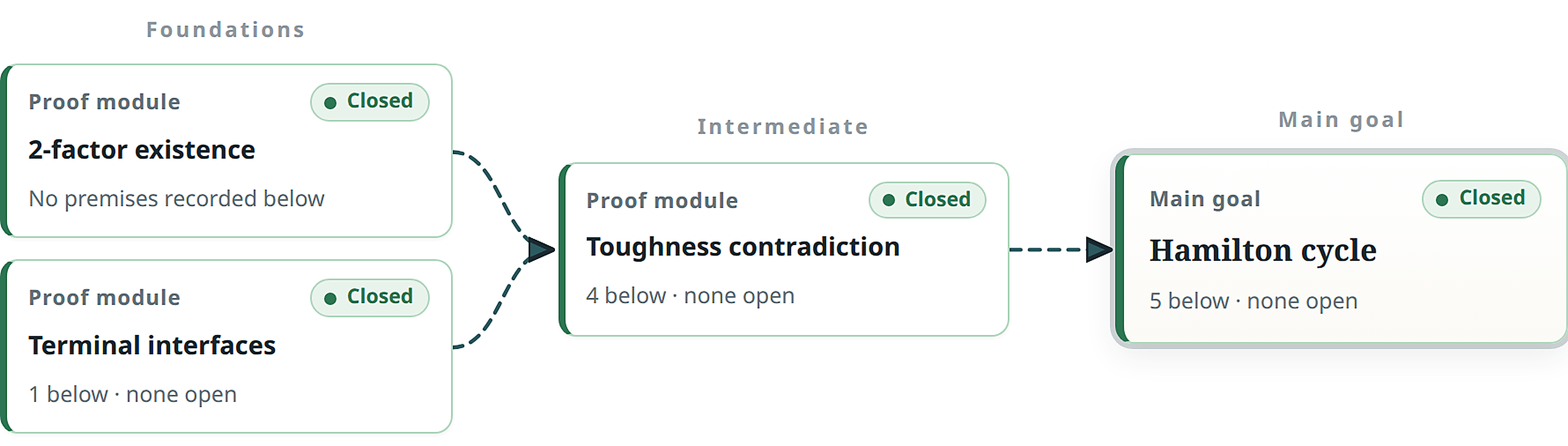}
\caption{Selected proof modules leading to Nikoghosyan's Hamiltonicity theorem. The overview exposes the principal prerequisite dependencies.}
\label{fig:case-niko}
\end{figure}

\Cref{fig:case-niko} shows a compact representation of the main proof modules. Factor existence and terminal interfaces support the toughness contradiction, which in turn leads to the Hamilton-cycle claim. This dependency structure makes the prerequisite chain available for inspection from the main goal.

\begin{figure}[!htbp]\centering
\includegraphics[width=0.78\textwidth]{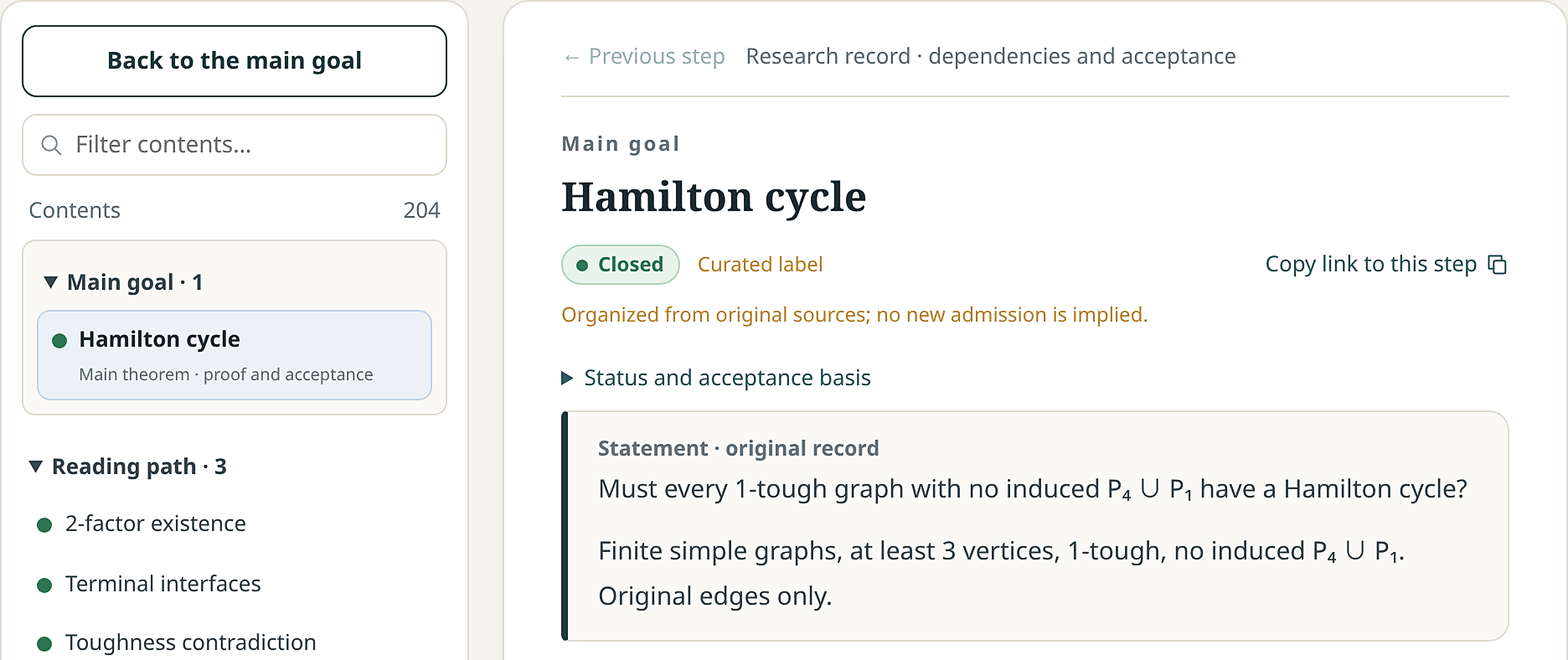}
\caption{Nikoghosyan's original target and its recorded acceptance basis in the proof reader.}
\label{fig:case-niko-reader}
\end{figure}

The corresponding proof-reader entry is shown in \cref{fig:case-niko-reader}. Selecting a module opens its source passage, premises, and recorded evidence. The historical main-goal display is associated with the archived formalization audit; the native Fact Graph preserves its separate admitted intermediate records.

\subsection{Lewis--Overton: comparing two exact asymptotic constructions}
\label{sec:bfgs}
Lewis and Overton investigated BFGS behavior for nonsmooth objective functions. Their 2008 analysis included a conjecture concerning the trial-normalized convergence rate of nonterminating runs on a tilted absolute-value objective \cite{lewisoverton2008}. The conjecture concerns the dependence of a sharp asymptotic rate on the tilt parameter and the initialization, under specified line-search and counting conventions. The related 2013 work provides broader context for nonsmooth quasi-Newton behavior \cite{lewisoverton2013}.

The archived investigation first required a precise formulation of the target. The source algorithm fixes the Armijo and weak Wolfe conditions, the treatment of trial step lengths and line-search failures, and the normalization applied to the asymptotic rate. Its counterexample search identified periodic patterns whose states return to scaled versions of their starting configurations. Exact rational calculations were used to investigate these cycles, followed by a parameterized argument establishing that two distinct mechanisms persist over an open interval.

For $9/5<u<13/6$, the two constructed execution families have cycle contractions
\begin{equation}
\label{eq:bfgs}
q_A(u)=\frac{3u}{4(u+1)^2},\qquad
q_B(u)=\frac{u}{(u+1)^2},
\qquad
\rho_A(u)=q_A(u)^{1/5},\quad
\rho_B(u)=q_B(u)^{1/4}<\rho_A(u).
\end{equation}
The five-trial and four-trial families therefore yield different sharp trial-normalized rates for the same fixed parameter. The corresponding manuscript and Lean development provide the detailed line-search calculations and interval conditions \cite{chenbfgs2026}.

\begin{figure}[!htbp]\centering
\begingroup
\setlength{\fboxsep}{0pt}
\fcolorbox{EulerFrameGray}{white}{%
\includegraphics[width=0.92\textwidth]{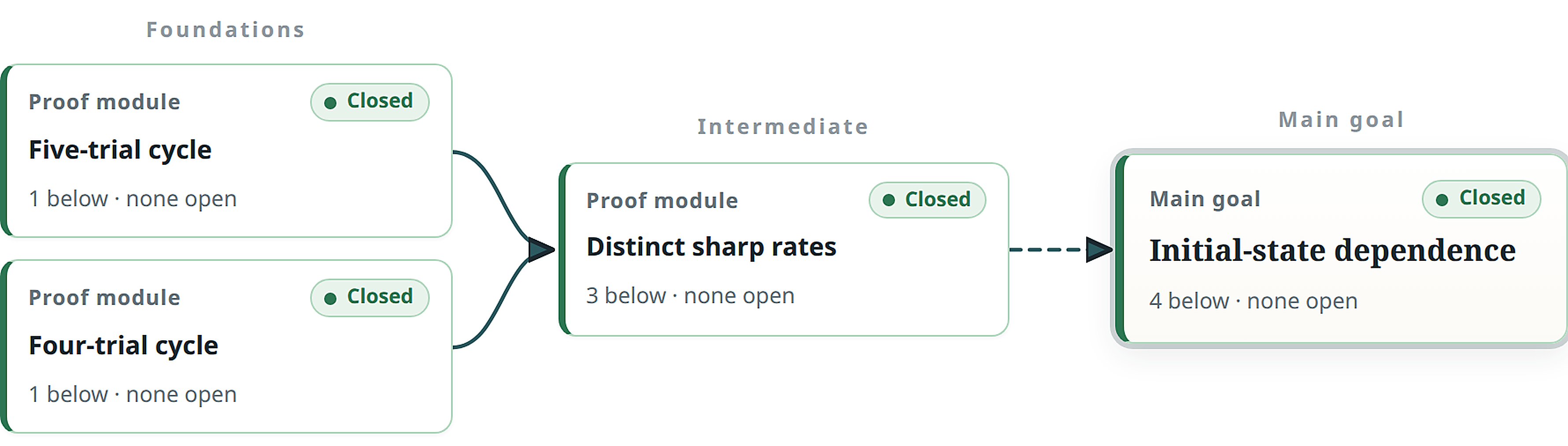}%
}
\endgroup
\caption{Proof structure for the Lewis--Overton investigation. The five-trial and four-trial execution families establish distinct sharp rates, which are compared to derive initialization dependence. Each module links to its exact conditions and evidence.}
\label{fig:case-bfgs}
\end{figure}

The Research Graph expresses the result through two independent branches. Each branch must establish its recurrence pattern, validity of line-search trials, and exact contraction. The branches meet at the rate comparison needed for the target conclusion. This proof structure is visible in \cref{fig:case-bfgs}. The workbench also retains earlier single-parameter experiments and the later interval-level obligation, allowing a reader to trace the move from candidate patterns to a general mathematical claim. A researcher can formulate a follow-up about extending one periodic mechanism by selecting that branch and specifying a new parameter scope.

\subsection{Jankiewicz--Norin--Wise: retaining constructive and negative research evidence}
\label{sec:jnw}
Jankiewicz, Norin, and Wise introduced legal systems as combinatorial certificates relevant to virtual fibering questions for right-angled Coxeter groups \cite{jnw2021}. Their Problem 5.2 asks for a graph satisfying global connectivity and curvature requirements while excluding a legal system in a specified family of subgraphs. We consider the four-connected version: a finite four-connected graph with girth at least four and nonnegative Charney--Davis curvature, such that every four-connected ordinary subgraph admits no legal system.

The companion result gives a graph with 33 vertices and 66 edges. It is four-regular and four-connected, has girth four and Charney--Davis curvature one, and satisfies the required legal-system obstruction \cite{chenjnw2026}. The formal development records the global graph properties and the ordinary-subgraph statement. Its proof combines structural arguments with finite certificates, some of which are checked through Lean's \code{native\_decide} procedure. The stronger variant involving all three-connected ordinary subgraphs remains a separate question.

The problem illustrates why unsuccessful candidate tests should remain part of the research state. Early exploration considered natural regular, connected, triangle-free graphs. Computation produced legal systems for several proposed witnesses. A positive legal-system certificate ruled out its particular candidate as a solution to the negative-witness task. These findings motivated a change from screening named graphs to constructing a host that preserves a known obstruction under a suitable modification. The later solution combines a seed obstruction, a four-cycle restriction argument, and a three-cap construction.

\begin{figure}[!htbp]\centering
\includegraphics[width=0.86\textwidth]{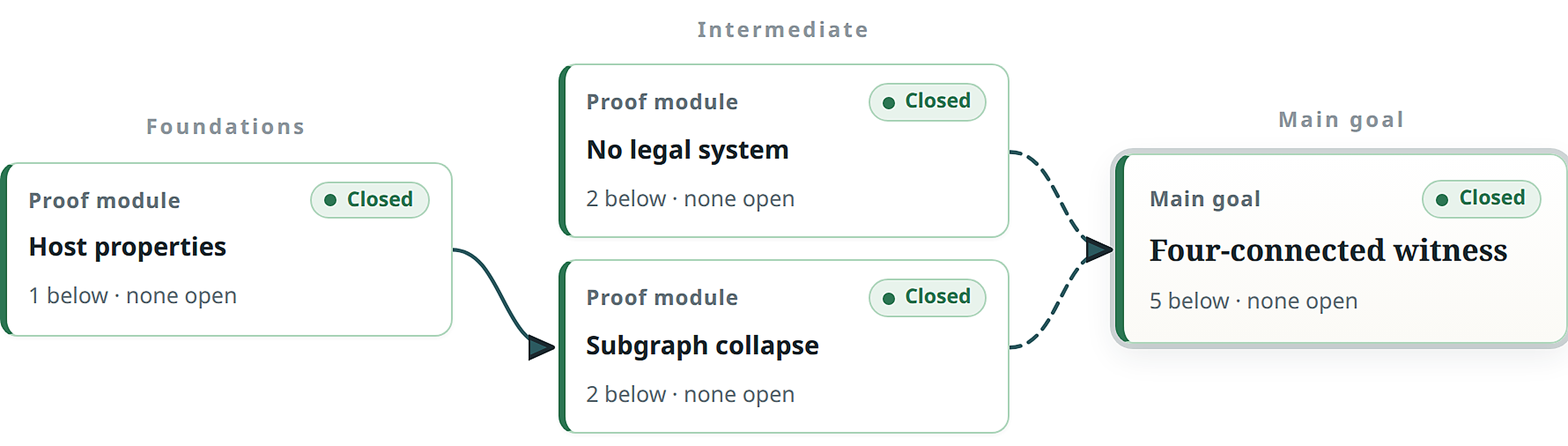}
\par\smallskip{\small (a) Global host properties and ordinary-subgraph obstruction.}\par\medskip
\begin{minipage}[t]{0.48\textwidth}\vspace{0pt}
\includegraphics[width=\linewidth]{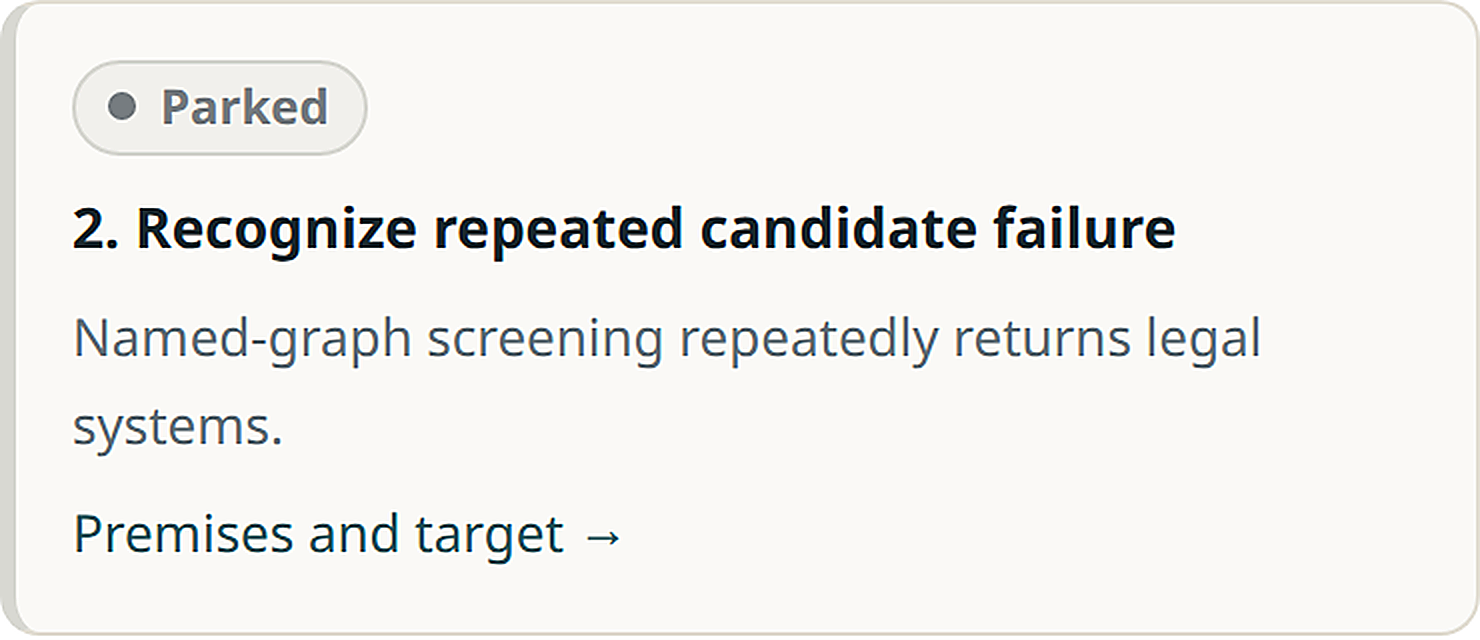}
\end{minipage}\hfill
\begin{minipage}[t]{0.48\textwidth}\vspace{0pt}
\includegraphics[width=\linewidth]{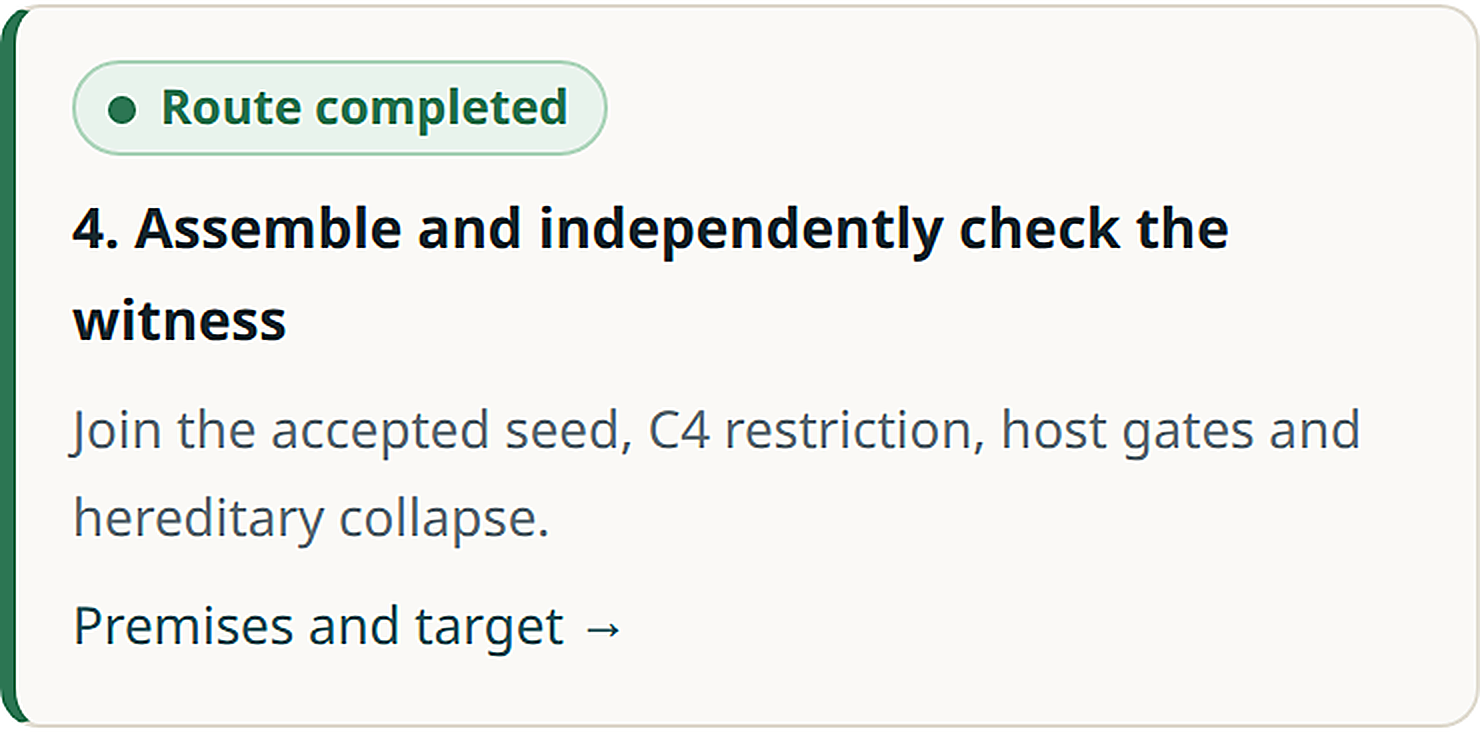}
\end{minipage}
\par\smallskip{\small (b) Retained unsuccessful search and completed construction route.}
\caption{The Jankiewicz--Norin--Wise case. The overview identifies the proof obligations of the constructed host, while the research-route records preserve the candidate-search history and the later successful construction mechanism.}
\label{fig:case-jnw}
\end{figure}

\Cref{fig:case-jnw}(a) summarizes two load-bearing requirements: establishing the host's global properties and showing the hereditary legal-system obstruction for its relevant subgraphs. Panel (b) presents a parked candidate-search route next to the completed constructive route. The record of an 11-vertex cap with a legal system also clarifies why this particular host does not satisfy the stronger three-connected requirement. These separate routes let a researcher inspect the outcome of each strategy and the mathematical scope of its evidence within one investigation.

\subsection{What the case studies show}
Across the three investigations, the workbench exposes different aspects of the mathematical process. The Nikoghosyan graph makes prerequisite completeness inspectable; Lewis--Overton retains the detailed semantics and competing constructions behind a rate comparison; Jankiewicz--Norin--Wise connects failed candidates to a change of construction strategy. Each case also provides an entry point for further inquiry: audit a prerequisite, vary a parameter condition, or revisit a parked route.

The companion Lean repositories and manuscripts provide the mathematical artifacts for each case. The visualizations preserve source-linked records of investigations conducted across earlier workflow versions, including accepted dependencies, unresolved steps, and changes of approach.

\section{Evaluation}
\label{sec:evaluation}
This section reports completed software and integration validation and describes the mathematical research records available for inspection in the three case-study archives. These results document the implemented workflow, its evidence boundaries, and the source material presented by the workbench.

\subsection{Software and integration validation}
\label{sec:validation}
At commit \code{ea505c7}, the repository's GitHub Actions workflow completed its offline tests on Python 3.10 and Python 3.13 with \textbf{1,352 passing tests and two skips} in each environment. The native-team workflow completed \textbf{56 tests on Linux} and \textbf{53 passes with three skips on macOS}. These runs covered protocol adapters and real operating-system isolation while using synthetic CLI sessions rather than paid model calls. A separate release workflow built the wheel and exercised the packaged dashboard in a Chromium browser, including all six workbench views, graph selection, language switching, an empty project, responsive layout, and project isolation checks. The CI configurations and current results are available through the project's Actions history.

Tests within the research-control package cover mathematical object schemas, scoped evidence, stable identities, event replay, resource reservations, scheduling decisions, context compilation, reconciliation, recurrence, formal-kernel adapter behavior, and final acceptance. Workspace and provider tests cover session preparation, checkpoint capture, restored branch identity, preservation of private files, and supported operating-system isolation. The use of model and checker fixtures permits deterministic examination of protocol behavior, including retry, stale-source, and failure cases.

\begin{table}[htbp]\centering\small
\caption{Validation artifacts associated with the current repository snapshot.}
\label{tab:validation}
\begin{tabularx}{\textwidth}{@{}L{0.27\textwidth}Y L{0.20\textwidth}@{}}\toprule
Component & Validation coverage & Observed CI outcome\\\midrule
Offline Python suite & Research state, scheduler, verification adapters, recovery, API, and evaluation utilities & 1,352 pass; 2 skip (both Python versions)\\
Native team protocols & Provider session doubles and filesystem isolation on Linux/macOS & 56 pass (Linux); 53 pass, 3 skip (macOS)\\
Packaged dashboard & Wheel installation, six views, source graph, i18n, small-screen layout & Browser job passed\\
Mathematical examples & Manuscripts, formal repositories, historical graph snapshots & Companion artifacts; see \cref{sec:cases}\\\bottomrule
\end{tabularx}
\end{table}

The human-control demonstration exercises selection of a canonical object, preview and confirmation, one focused task, and return of an evidence-tagged outcome. Deterministic worker and verifier responses travel through the implemented HTTP, dispatch, lease, gateway, and reconciliation services. The demo retains the original parent objective while recording each derived investigation. A source-linked timeline makes the sequence inspectable in the browser (\cref{fig:intervention-timeline}). A separate two-provider restoration exercise used Codex and Grok on the infinitude-of-primes exercise. That exercise tested native conversation forking, checkpoint file consistency, and renewed lead--assistant exchange in the restored workspace.

These results verify the software's state transitions, provider protocol handling, and workbench behavior under the reported test configurations.

\subsection{Mathematical trace inspection}
The three case-study archives contain inspectable, source-linked research records. The Nikoghosyan view exposes the prerequisite structure behind the Hamiltonicity statement; the Lewis--Overton view preserves the two execution branches and their sharp-rate comparison; the Jankiewicz--Norin--Wise view records the construction route alongside earlier candidate exclusions. These examples are presented in \cref{sec:cases}.

The historical views retain original mathematical passages, graph relations, and available checking provenance. Their manuscripts and formalization repositories provide the corresponding mathematical artifacts.

\section{Related Work}
\label{sec:related}
\paragraph{Mathematical reasoning agents.}
Danus combines multiple reasoning agents with Fact-Graph memory and verification to support sustained mathematical exploration \cite{danus2026}. \Euler builds on that admission and memory foundation and extends the research-control and human-workbench layers. Rethlas/Archon investigates automated conjecture resolution with formal checking \cite{rethlas2026}; MMAT coordinates mathematical agents through control and execution roles \cite{mmat2026}. Albilich explores steerable proof-state orchestration and computer algebra, while Magenta and LEAP develop workflows with formal verification feedback \cite{albilich2026,magenta2026,leap2026}. These systems provide complementary designs for orchestrating extended mathematical work.

\paragraph{Human--AI mathematical collaboration.}
AI Co-Mathematician studies a workbench organized around mathematicians' participation, progressive reading of research artifacts, and steering of investigations \cite{aicomath2026}. Human feedback and intervention also appear in multi-agent research systems such as Danus and MMAT \cite{danus2026,mmat2026}. \Euler places a canonical mathematical object at the center of this interaction: the object read by the researcher is the same object referenced by the scheduler, context compiler, and reconciled outcome. Milestones extend this continuity across saved agent-team states.

\paragraph{Scientific exploration and persistent context.}
AI Co-Scientist and AI Scientist-v2 organize iterative scientific exploration through hypothesis management, feedback, and search \cite{coscientist2025,aiscientistv22025}. Work on long-context access and agentic context engineering motivates structured approaches to recalling relevant information from large histories \cite{lostmiddle2024,ace2025}. \Euler specializes these ideas to mathematical premises, obligations, counterexamples, and proof dependencies, using the current research frontier to determine which historical material enters each agent task.

\paragraph{Neural and symbolic representations.}
Neural--symbolic research considers ways of combining learned methods with explicitly represented knowledge and constraints \cite{yangnesy2025,wangnesy2025}. The symbolic structures in \Euler specify research objects, their identities, relations, and permissible evidence-supported transitions. Language models provide proof strategies and mathematical arguments within this persistent scaffold, while theorem checkers and exact computations supply domain-specific validation.

\section{Conclusion and Future Directions}
\label{sec:discussion}
The central design principle of \Euler is that mathematical research should remain intelligible and actionable throughout an extended investigation. A shared representation allows the system to coordinate agent work while giving researchers direct access to the mathematical structures underlying that work. Exact statements and scopes anchor the investigation; proof routes and blockers expose its developing logic; evidence records explain which conclusions have been admitted; and milestones preserve its history. The active frontier connects these records to the next actions the system can perform. The implemented workbench makes the recorded proof dependencies and research decisions inspectable, while the control plane connects focused human questions and agent actions to their context, verification evidence, and resulting updates. The three mathematical cases illustrate these capabilities in concrete investigations.

Several research questions remain. The usefulness of the representation depends on the quality of the proposed routes, the accuracy of scope assignments, and the ability to retrieve the relevant dependencies and failures. LLM proof review has a different assurance level from Lean-kernel verification, and formal statements require alignment with their intended mathematical meaning. Applying the current architecture to a wider range of mathematical fields and problem classes will help determine its generality, reliability, and practical value for sustained reasoning and human--AI mathematical collaboration.

\section*{Acknowledgements}
Euler's graph design draws on \href{https://github.com/frenzymath/Danus}{Danus}, and we consulted the \href{https://github.com/frenzymath/Rethlas}{Rethlas} implementation during development. We thank the authors and contributors of both projects for sharing their work.

We thank Professor Michael Overton and Professor Adrian Lewis for the time and care they devoted to reviewing our argument concerning the Lewis--Overton conjecture.

We also thank the mathematicians and research communities whose prior results, definitions, and lines of reasoning made this work possible. Euler's use of mathematical knowledge and proof structure depends fundamentally on human insight.

\appendix
\small
\section{Implementation and Record Interfaces}
\label{app:records}
An event in the research-control store includes the project identifier, actor, event type, idempotency key, timestamp, payload hash, and optional causation and correlation identifiers. The control plane replays these events to recover contract, obligation, route, budget, and execution states. Immutable context and execution receipts connect the selected action to its worker result.

A focused human request additionally identifies the source object and reviewed target. Representative fields include \code{object\_type}, \code{object\_id}, \code{object\_version}, \code{statement\_hash}, \code{scope\_hash}, \code{contract\_id}, \code{intent}, \code{user\_question}, \code{requested\_statement}, \code{requested\_scope}, and resource-policy information. The current object and research revision are checked when the request is admitted and again when its action starts. The response timeline records the linked derived objects, task lease, evidence, and reconciliation outcome.

\section{Reproducibility and Artifact Index}
\label{app:artifacts}
The evaluated Euler source is version \code{0.3.0.dev0}, revision
\href{https://github.com/canghaimeng/Euler/commit/ea505c71bb846080728df722e541175d4495cd02}{\code{ea505c7}}
(October 13, 2026). Code, test workflows, and documentation are available at
\url{https://github.com/canghaimeng/Euler}. The mathematical case materials are under
\path{examples/nikoghosyan}, \path{examples/lewis-overton}, and \path{examples/jnw},
with linked paper PDFs and Lean developments. \path{docs/DASHBOARD-GUIDE.md} and
\path{docs/TEAM-RECOVERY.md} describe the personal research workflow.


\begin{thebibliography}{99}
\raggedright

\bibitem{rethlas2026}
Haocheng Ju, Guoxiong Gao, Jiedong Jiang, Bin Wu, Zeming Sun, Shurui Liu,
Leheng Chen, Yutong Wang, Yuefeng Wang, Zichen Wang, Wanyi He, Peihao Wu,
Liang Xiao, Ruochuan Liu, Bryan Dai, and Bin Dong.
\newblock Automated Conjecture Resolution with Formal Verification.
\newblock \arxivref{2604.03789}, 2026.

\bibitem{danus2026}
Jihao Liu, Guoxiong Gao, Zeming Sun, Bin Wu, Shurui Liu, Jiedong Jiang,
Haocheng Ju, Leheng Chen, Ronnie Cheng, Xiping Zhang, and Bin Dong.
\newblock Danus: Orchestrating Mathematical Reasoning Agents with Fact-Graph Memory.
\newblock \arxivref{2607.06447}, 2026.

\bibitem{aicomath2026}
Daniel Zheng, Ingrid von Glehn, Yori Zwols, Iuliya Beloshapka, Lars Buesing,
Daniel M. Roy, Martin Wattenberg, Bogdan Georgiev, Tatiana Schmidt, Andrew Cowie,
et al.
\newblock AI Co-Mathematician: Accelerating Mathematicians with Agentic AI.
\newblock \arxivref{2605.06651}, 2026.

\bibitem{mmat2026}
Yichuan Cao, Ruichen Qiu, Junqi Liu, Jiaqi Wang, Dakai Guo, Ruyong Feng,
Lihong Zhi, and Xiao-Shan Gao.
\newblock MechMath Agent Team: LLM Driven Agents for Mathematical Research.
\newblock \arxivref{2607.04394}, 2026.

\bibitem{lostmiddle2024}
Nelson F. Liu, Kevin Lin, John Hewitt, Ashwin Paranjape,
Michele Bevilacqua, Fabio Petroni, and Percy Liang.
\newblock Lost in the middle: How language models use long contexts.
\newblock \emph{Transactions of the Association for Computational Linguistics},
12:157--173, 2024.
\newblock \href{https://doi.org/10.1162/tacl_a_00638}{doi:10.1162/tacl\_a\_00638}.

\bibitem{ace2025}
Qizheng Zhang, Changran Hu, Shubhangi Upasani, Boyuan Ma, Fenglu Hong,
Vamsidhar Kamanuru, Jay Rainton, Chen Wu, Mengmeng Ji, Hanchen Li,
Urmish Thakker, James Zou, and Kunle Olukotun.
\newblock Agentic context engineering: Evolving contexts for self-improving
language models.
\newblock \arxivref{2510.04618}, 2025; revised 2026.

\bibitem{chenbfgs2026}
Qiuyu Chen.
\newblock Initialization-dependent BFGS trial rates for tilted absolute values.
\newblock \arxivref{2609.09872}, 2026.

\bibitem{chenjnw2026}
Qiuyu Chen.
\newblock A four-connected graph without a legal system.
\newblock \arxivref{2609.13419}, 2026.

\bibitem{nikoghosyan2013}
Zh.~G. Nikoghosyan.
\newblock Disconnected forbidden subgraphs, toughness, and Hamilton cycles.
\newblock \emph{ISRN Combinatorics}, 2013, Article 673971.
\newblock \href{https://doi.org/10.1155/2013/673971}{doi:10.1155/2013/673971}.
\newblock Preprint: \arxivref{1207.5132}.

\bibitem{shan2025}
Songling Shan.
\newblock Hamiltonian cycles in tough $(P_4\cup P_1)$-free graphs.
\newblock \arxivref{2504.08936v2}, 2025.

\bibitem{chennikoghosyan2026}
Qiuyu Chen.
\newblock Nikoghosyan's conjecture: Lean 4 formalization.
\newblock Software artifact, 2026; revision \code{69f3a96}.
\newblock \url{https://github.com/canghaimeng/Nikoghosyan-conjecture}.

\bibitem{eulersoftware2026}
Qiuyu Chen.
\newblock Euler: Towards Human-Centered Multi-Agent Mathematical Discovery.
\newblock Software snapshot \code{0.3.0.dev0}, revision \code{ea505c7}, October 13, 2026.
\newblock \url{https://github.com/canghaimeng/Euler}.

\bibitem{lewisoverton2008}
Adrian S. Lewis and Michael L. Overton.
\newblock Nonsmooth optimization via BFGS.
\newblock Optimization Online technical report, 2008.
\newblock \url{https://optimization-online.org/2008/12/2172/}.

\bibitem{lewisoverton2013}
Adrian S. Lewis and Michael L. Overton.
\newblock Nonsmooth optimization via quasi-Newton methods.
\newblock \emph{Mathematical Programming}, 141:135--163, 2013.
\newblock \href{https://doi.org/10.1007/s10107-012-0514-2}{doi:10.1007/s10107-012-0514-2}.

\bibitem{jnw2021}
Kasia Jankiewicz, Sergey Norin, and Daniel T. Wise.
\newblock Virtually fibering right-angled Coxeter groups.
\newblock \emph{Journal of the Institute of Mathematics of Jussieu},
20(3):957--987, 2021.
\newblock \href{https://doi.org/10.1017/S1474748019000422}{doi:10.1017/S1474748019000422}.

\bibitem{albilich2026}
Ting Gong, Michael Ruofan Zeng, and Yong Yang.
\newblock Albilich: Steerable Proof-State Orchestration for LLM-Based Mathematical Research with CAS Integration.
\newblock \arxivref{2607.27705}, 2026.

\bibitem{magenta2026}
Joshua Ong Jun Leang, Haonan Li, Zheng Zhao, Xinyi Shang, Wenda Li,
Zhengzhong Liu, Erix Xing, Shay Cohen, and Eleonora Giunchiglia.
\newblock Magenta: Closing the Loop Between Mathematical Reasoning and Lean Verification.
\newblock \arxivref{2609.11319}, 2026.

\bibitem{leap2026}
Po-Nien Kung, Linfeng Song, Dawsen Hwang, Jinsung Yoon, Chun-Liang Li,
Simone Severini, Mirek Ol\v{s}\'ak, Edward Lockhart, Quoc V. Le, Burak Gokturk,
Thang Luong, Tomas Pfister, and Nanyun Peng.
\newblock LEAP: Supercharging LLMs for Formal Mathematics with Agentic Frameworks.
\newblock \arxivref{2606.03303}, 2026.

\bibitem{coscientist2025}
Juraj Gottweis, Wei-Hung Weng, Alexander Daryin, et al.
\newblock Towards an AI co-scientist.
\newblock \arxivref{2502.18864v1}, 2025.

\bibitem{aiscientistv22025}
Yutaro Yamada, Robert Tjarko Lange, Cong Lu, Shengran Hu, Chris Lu,
Jakob Foerster, Jeff Clune, and David Ha.
\newblock The AI Scientist-v2: Workshop-level automated scientific discovery
via agentic tree search.
\newblock \arxivref{2504.08066}, 2025.


\bibitem{yangnesy2025}
Xiao-Wen Yang, Jie-Jing Shao, Lan-Zhe Guo, Bo-Wen Zhang, Zhi Zhou,
Lin-Han Jia, Wang-Zhou Dai, and Yu-Feng Li.
\newblock Neuro-Symbolic Artificial Intelligence: Towards Improving the Reasoning Abilities of Large Language Models.
\newblock In \emph{Proceedings of IJCAI}, Survey Track, pages 10770--10778, 2025.
\newblock \url{https://doi.org/10.24963/ijcai.2025/1195}.

\bibitem{wangnesy2025}
Wenguan Wang, Yi Yang, and Fei Wu.
\newblock Towards Data-And Knowledge-Driven AI: A Survey on Neuro-Symbolic Computing.
\newblock \emph{IEEE Transactions on Pattern Analysis and Machine Intelligence},
47(2):878--899, 2025.
\newblock \url{https://doi.org/10.1109/TPAMI.2024.3483273}.
\end{thebibliography}
\end{document}